\documentclass[letterpaper]{article} 
\usepackage{aaai2027} 
\usepackage[hyphens]{url}  
\usepackage{graphicx} 
\usepackage{natbib}  
\usepackage{caption} 
\usepackage{multirow}
\usepackage{cite}

\usepackage{adjustbox}

\usepackage{graphicx}
\usepackage{textcomp}
\usepackage{makecell}
\usepackage{color}

\usepackage{amssymb}
\usepackage{algorithm}    
\usepackage{algpseudocode}  
\usepackage{amsmath}        
\usepackage{amssymb}        
\usepackage{booktabs}  

\usepackage{xcolor}

\definecolor{fgred}{HTML}{D93123}
\definecolor{bgyellow}{HTML}{F5C945}
\definecolor{pblue}{HTML}{4FADEA}

\usepackage{newfloat}
\usepackage{listings}
\DeclareCaptionStyle{ruled}{labelfont=normalfont,labelsep=colon,strut=off} 
\floatstyle{ruled}
\newfloat{listing}{tb}{lst}{}
\floatname{listing}{Listing}

\usepackage[table]{xcolor} \definecolor{lightgreen}
{HTML}{F0FCF0}

\usepackage[table]{xcolor} \definecolor{lightblue}
{HTML}{F0FCF0}

\usepackage[table]{xcolor} \definecolor{lightgreen}
{HTML}{F0FCF0}

\usepackage{multirow}   
\usepackage{booktabs}   
\usepackage{graphicx}   
\usepackage{colortbl}   

\usepackage{booktabs}

\title{NeoRed: A Knowledge-Logic-Alignment MLLM for Neonatal Respiratory Disease Diagnosis}

\author{
    Yinan Liu\textsuperscript{1},
    Hongtai Xia\textsuperscript{1},
    Haoran Xu\textsuperscript{1},
    Jiankang Hong\textsuperscript{1},
    Yu Jianli\textsuperscript{2},
    Jingkuan Song\textsuperscript{1},
    Ye Luo\textsuperscript{1}\corresponding
}

\affiliations{
    \textsuperscript{1}School of Computer Science and Technology, Tongji University, Shanghai, China\\
    \textsuperscript{2}Department of Radiology, Shanghai First Maternity and Infant Hospital, Shanghai, China
}

\begin{document}

\maketitle

\begin{abstract}
Neonatal respiratory diseases are a major cause of neonatal morbidity and mortality, posing substantial challenges in clinical practice. Despite recent advances, existing Multimodal Large Language Models (MLLMs) face two key limitations in neonatal diagnosis: (1) domain gap arising from predominantly adult training data; (2) insufficient integration of multidimensional clinical context for accurate diagnosis. To address these challenges, we collect two real-world clinical datasets (NeoCXR and NeoCXR-EV) and propose NeoRed, to the best of our knowledge, the first MLLM tailored for neonatal respiratory disease, filling the gap in neonatal diagnostic reports generation. To enhance joint diagnosis from heterogeneous clinical context and chest X-rays, we design a novel Knowledge–Logic–Alignment (KLA) framework which constrains model behavior from three perspectives: 1) Knowledge Prior Injection (KPI) incorporates neonatologist-inspired diagnostic priors into multimodal representations, guiding disease-specific attention across modalities; 2) Diagnostic Logic Constraint (DLC) aligns the semantics of generated reports with multimodal diagnostic logic; and 3) Visual Semantic Alignment (VSA) establishes semantic correspondence between visual features and imaging conclusions. Extensive experiments demonstrate that NeoRed enables accurate neonatal diagnostic reports generation, achieving ROUGE-L of 53.29\% and Clinical Efficacy F1 score of 65.19\% on NeoCXR, outperforming existing MLLMs. NeoRed also preserves competitive report generation performance on adult benchmarks (MIMIC-CXR and IU-Xray). Datasets will be available upon application.
\end{abstract}



%

\section{Introduction}

Neonatal respiratory diseases, including Neonatal Respiratory Distress Syndrome (NRDS), Transient Tachypnea of the Newborn (TTN), and neonatal pneumonia, represent one of the leading causes of morbidity and mortality among newborns globally ~\cite{Neo1,Neo2}. Early and accurate diagnosis is pivotal for timely intervention and improving clinical outcomes \cite{neo_nomogram,neocxr}. 



\begin{figure}[htp]
    \centering
    \includegraphics[width=\linewidth]{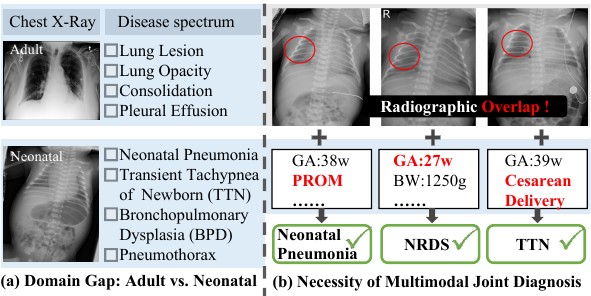}
    \vspace{-1.5em}
    \caption{Motivations of the proposed NeoRed model. (a) Domain gaps in Chest X-ray and disease spectrum between neonatal and adult. (b) Necessity of integrating CXRs with clinical context for accurate neonatal diagnosis.}
    \label{fig:f1}
    \vspace{-1.2em}
\end{figure}

In recent years, Multimodal Large Language Models (MLLMs) have shown strong capabilities in joint vision–language understanding and reasoning, achieving notable progress in tasks such as image captioning \cite{clip,blip2} and multimodal dialogue \cite{Dialog,Dialogshikra}. Inspired by these advancements, recent studies have explored adapting MLLMs to medical domain, particularly for tasks such as radiology report generation \cite{vlmrg,medclip} and medical visual question answering \cite{pefomed,medvqa2}. 
However, despite these promising advances, their application to neonatal clinical scenarios remains challenging due to two main reasons. \textbf{First, as shown in Fig.~\ref{fig:f1}(a), neonatal and adult populations exhibit substantial domain gaps in both radiographic appearance and disease spectrum.} Neonatal CXRs show immature anatomy and smaller, lower-contrast lesions, while neonatal diseases such as NRDS, TTN, and BPD differ markedly from common adult conditions. Existing MLLMs are primarily trained on adult data. Given the severe domain shift in both visual features and disease spectrum, directly transferring such models to neonatal CXRs yields limited effectiveness. \textbf{Second, neonatal respiratory diseases often exhibit substantial radiographic overlap, making them difficult to distinguish on CXRs.} As shown in Fig.~\ref{fig:f1}(b), neonatal pneumonia, NRDS, and TTN can present with highly similar radiographic patterns in the same lung regions, limiting diagnosis based on imaging alone. However, incorporating clinical information enables more accurate differentiation. For example, the first case can be identified as pneumonia when combined with premature rupture of membranes; the second as NRDS when considered alongside prematurity and extremely low birth weight; and the third as TTN when cesarean delivery is taken into account. In clinical practice, neonatologists therefore integrate multiple sources of clinical information to support comprehensive diagnosis and reduce misdiagnosis. while existing MLLMs primarily focus on vision–language alignment and natural language generation. Due to lack of explicit training constraints, such models fail to prioritize key clinical indicators, limiting joint diagnosis from CXR and clinical context.


To address these limitations, we propose the Knowledge-Logic Alignment Multimodal Large Language Model for \textbf{Neo}natal \textbf{Res}piratory Disease \textbf{D}iagnosis (\textbf{NeoRed}). \textit{To the best of our knowledge, NeoRed is the first MLLM tailored for neonatal respiratory disease diagnosis, filling the gap in neonatal radiology report generation.} NeoRed jointly incorporates neonatal CXRs and clinical context as model inputs to generate reports containing imaging conclusions and disease diagnosis, enabling multimodal diagnosis for neonatal respiratory disease. Specifically, to address the domain gap between adult and neonatal populations and the scarcity of neonatal data, we collaborate with two partner hospitals to curate two real-world multimodal datasets: NeoCXR and NeoCXR External Validation (NeoCXR-EV). Together, NeoCXR and NeoCXR-EV form a dual-center benchmark for training and evaluating MLLMs for neonatal respiratory disease diagnosis. To address the limited capability of existing models in jointly diagnosis over heterogeneous clinical information, we design a Knowledge–Logic– Alignment (KLA) framework that inspired by the multimodal diagnostic workflow of neonatologists. Specifically, KLA constrains model behavior from three perspectives: 1) Knowledge Prior Injection (KPI), which injects neonatologist-inspired diagnostic priors into multimodal representations for disease-specific attention; 2) Diagnostic Logic Constraint (DLC), which enforces semantic–diagnostic consistency by aligning global diagnostic-semantic anchor of report generation with the diagnostic logic of neonatologist; 3) Visual Semantic Alignment (VSA), which establishes bidirectional semantic correspondence between image features and conclusions via cross-modal contrastive learning, aligning visual evidence with its clinical interpretation. By coordinating multimodal clinical information in accordance with neonatologists’ diagnostic workflow, the proposed KLA enables accurate diagnosis of neonatal respiratory diseases. In summary, the key contributions of this paper are fourfold:

\begin{itemize}
    \item We construct NeoCXR and NeoCXR-EV, two real-world multimodal neonatal report generation datasets that fill a critical domain gap and will be released through application-based access to support future research.
    \item We propose NeoRed, to the best of our knowledge, the first MLLM for neonatal respiratory disease diagnosis, which jointly interprets chest X-rays and clinical context to generate accurate neonatal diagnostic reports.
    \item We propose a novel KLA framework to enhance multimodal joint diagnosis, where KPI injects expert priors into multimodal representations, DLC enforces semantic–diagnostic consistency, and VSA aligns imaging conclusions with visual evidence.
    \item Extensive experiments demonstrate that NeoRed outperforms 8 mainstream MLLMs on neonatal benchmarks (NeoCXR, NeoCXR-EV) while remaining competitive on adult benchmarks (MIMIC-CXR, IU-Xray).
\end{itemize}



\section{Related Work}

\subsection{Multimodal Large Language Models}
General MLLMs have advanced cross-modal representation learning and vision–language alignment \cite{Qwen2.5-vl,zhipu,Intern2.5-vl,bai2023qwen,Llava-next}. Early models, including Flamingo \cite{Flamingo} and the BLIP series \cite{blip,instructblip,blip2}, established strong multimodal representations through large-scale image–text pretraining. Subsequent models such as LLaVA \cite{llava-1.5}, Qwen-VL \cite{Qwen2.5-vl,Qwen3-vl}, and InternVL \cite{Intern2.5-vl} project visual features into the language space and adopt instruction tuning for multimodal interaction. LLaVA-NeXT \cite{Llava-next} and LLaVA-OneVision \cite{Llava-onevision} further improve visual alignment and reasoning through enhanced visual encoding and multi-stage training. However, their limited domain-specific medical knowledge constrains their reliability in medical diagnosis.

\subsection{Medical Multimodal Large Language Models}
Recent studies have extended MLLMs to radiology report generation and medical visual question answering \cite{Huatuogpt,Llava-med,LLaVA-rad,llava-ultra,Lingshu,medplib}. General medical models, including BiomedGPT \cite{BiomedGPT}, LLaVA-Med \cite{Llava-med}, UMIT \cite{Umit}, HuatuoGPT-Vision \cite{Huatuogpt}, and Lingshu \cite{Lingshu}, use large-scale multi-stage training to enhance multimodal alignment and diagnostic reasoning. Task-specific models such as LLaVA-Ultra \cite{llava-ultra}, LLaVA-Rad \cite{LLaVA-rad}, and RadFM \cite{RadFM} further improve radiological understanding across targeted 2D and 3D settings. However, most existing medical MLLMs are trained primarily on adult data and remain limited in modeling neonatal-specific disease patterns and diagnostic processes.


\section{NeoCXR and NeoCXR-EV Datasets}
\subsection{Data Collection and Process}
\begin{figure}[h]
  \centering
  \includegraphics[width=1.0\linewidth]{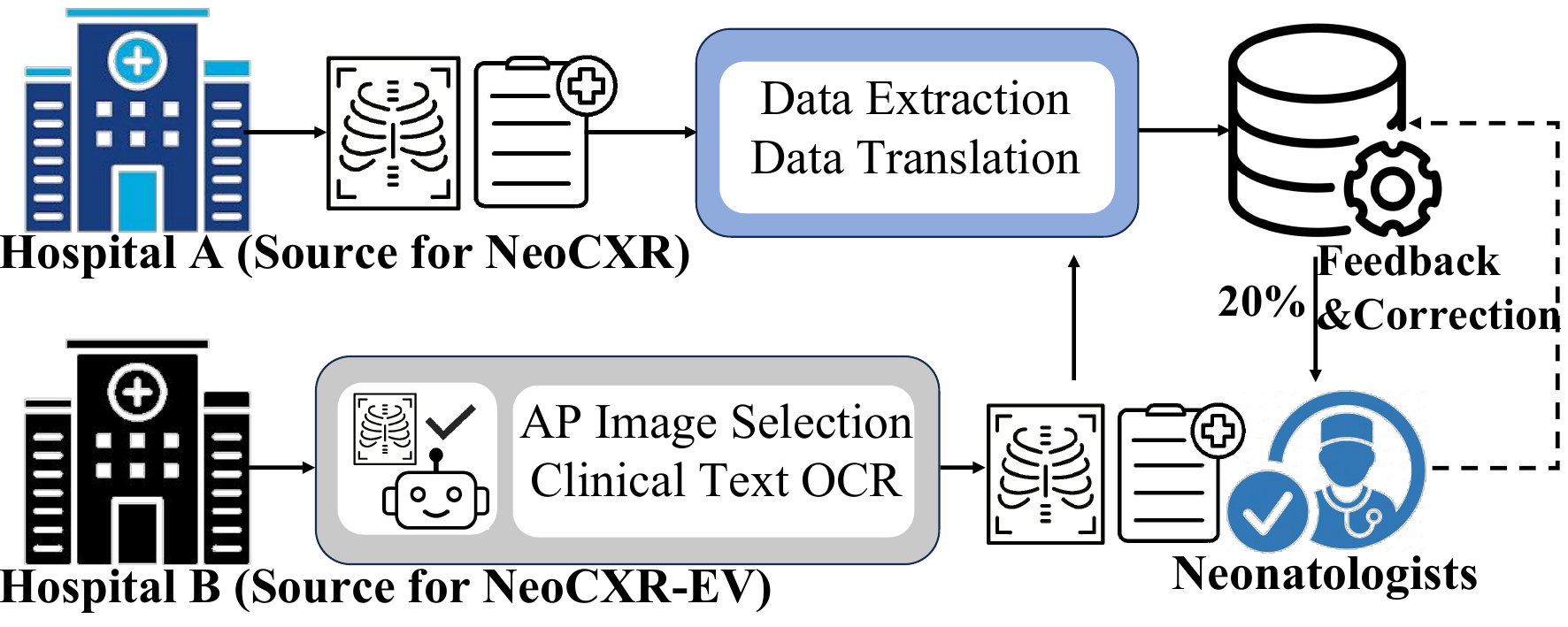}
  \vspace{-1.8em}
  \caption{Data collection and process pipeline of NeoCXR and NeoCXR-EV. AP denotes anteroposterior projection.}
  \label{fig:datapipline}
  \vspace{-0.7em}
\end{figure}

To support precise diagnosis of neonatal respiratory diseases, we construct NeoCXR and NeoCXR-EV datasets, which are collected from two independent hospitals. We designed a heterogeneous data pipeline to handle the distinct data from each hospital, as shown in Fig.~\ref{fig:datapipline}. Hospital A provided ready-to-use Anteroposterior (AP) CXRs and metadata, from which we directly extract the relevant data. While hospital B provided raw data, including multi-view CXRs and PDF-formatted clinical records. 
We first employed Zhipu GLM-4-Flash \cite{zhipu} to automatically identify AP views from multi-view images. Text was first extracted from the PDF documents using PaddleOCR~\cite{paddleocr}, followed by extraction of structured metadata. Then data from both hospitals were structured and translated into English by Tencent Cloud API. To ensure data quality, we performed iterative quality control. In each round, a new random 20\% subset of each dataset was independently reviewed by two neonatologists for AP-view identification, OCR, translation, and disease annotation. Errors were corrected based on expert feedback, and the process was repeated until no further errors were identified.

\begin{table}[ht]
\centering
\fontsize{8pt}{8pt}\selectfont
\renewcommand{\arraystretch}{0.66}
\setlength{\tabcolsep}{1.5pt}
\begin{tabular*}{0.98\columnwidth}{
@{\extracolsep{\fill}}lccccc@{}
}
\toprule
Dataset & Patients & Samples & Train & Val & Test \\
\midrule
NeoCXR    & 2,466 & 6,278 & 4,219 & 625 & 1,434 \\
NeoCXR-EV & 590   & 1,089 & --    & --  & 1,089 \\
\bottomrule
\end{tabular*}
\vspace{-0.8em}
\caption{Summary of the NeoCXR dataset series.}
\label{tab:datasets}
\vspace{-2em}
\end{table}

We finally obtain two datasets. NeoCXR, collected from Hospital A, contains 6,278 samples from 2,466 patients and is split at the patient level into training, validation, and internal test sets with a ratio of 7:1:2. NeoCXR-EV, collected from Hospital B, includes 1,089 samples from 590 patients and serves as an external validation set. Details are summarized in Tab.~\ref{tab:datasets}. Both datasets were approved by the relevant institutional ethics committees, and all patient data were de-identified before analysis. Each sample in the datasets is a multimodal tuple: the inputs consist of a chest X-ray paired with 21 clinical factors, while the outputs are structured reports comprising both imaging conclusion and disease diagnosis. These datasets cover 7 neonatal respiratory diseases (NRDS, neonatal pneumonia, TTN, pneumothorax, BPD, pleural effusion, atelectasis) and a normal category. Disease distribution is provided in \textit{supplementary material}.
\subsection{Structured Clinical Context}
To better model the clinical diagnostic logic, we structure complex clinical factors into three categories, according to the causal progression of neonatal diseases and the diagnostic logic of neonatologists. The factors are organized as follows:

\textbf{1) Developmental factors}, which capture neonatal maturity and growth, including gestational age, birth weight, body length, head circumference, multiple pregnancy, fetal growth restriction (FGR) or small for gestational age (SGA).

\textbf{2) Perinatal risks}, which capture maternal and obstetric conditions that may adversely affect the neonate, including delivery mode, premature rupture of membranes (PROM), preeclampsia, gestational hypertension, gestational diabetes, antenatal dexamethasone and magnesium sulfate.

\begin{figure}[h]
  \centering
  \includegraphics[width=1.0\linewidth]{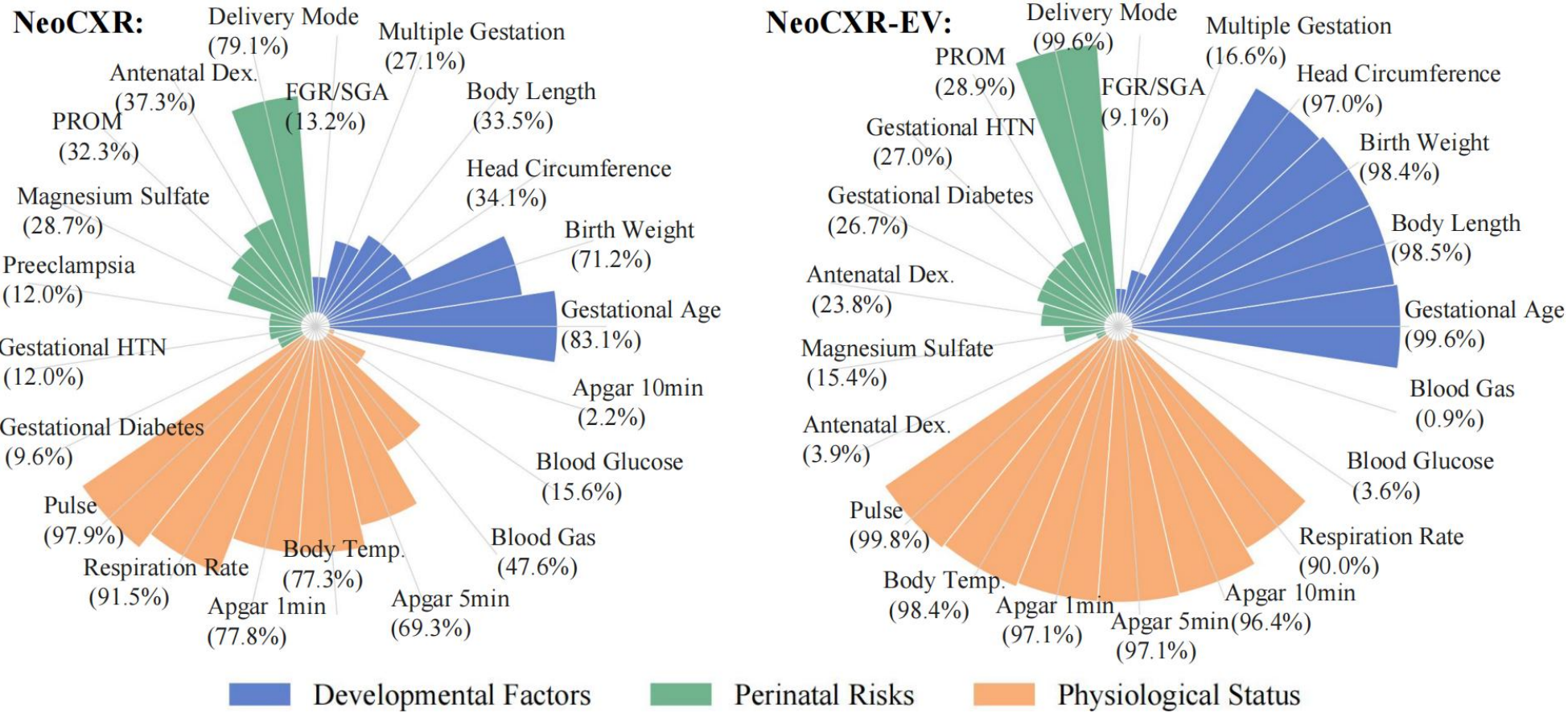}
    \vspace{-1.6em}
  \caption{Distribution of clinical context. Percentages indicate the completeness rate of each clinical factor.}
  \label{fig:clinicalcontext}
  \vspace{-1.5em}
\end{figure}

\textbf{3) Physiological status}, which represents the newborn's immediate postnatal condition, encompassing Apgar scores, body temperature, respiration rate, pulse, blood gas, and blood glucose. The distribution of clinical context on both datasets is shown in Fig.~\ref{fig:clinicalcontext}. Clinical factors are prevalent in both datasets, while notable distributional differences between NeoCXR and NeoCXR-EV.

To provide the model with explicit boundaries across clinical categories, each category is enclosed within relevant tokens: "<dev>...</dev>", "<peri>...</peri>", and "<phys>...</phys>". Missing categories are replaced with "not provided" for input consistency.

\section{Methodology}
\subsection{Overview of NeoRed}
\begin{figure*}[t]
  \centering
  \includegraphics[width=1.0\linewidth]{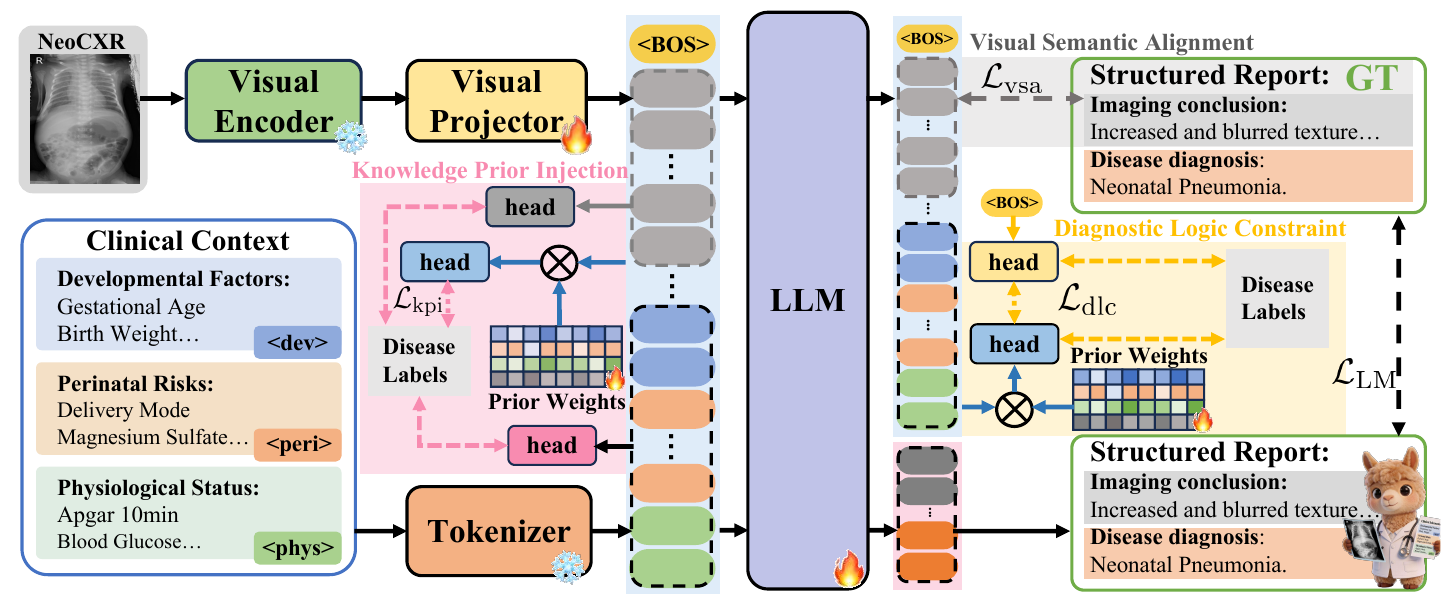}
  \vspace{-1.5em}
  \caption{The overview of NeoRed based on the proposed KLA framework. KLA consists of three modules: KPI, which injects neonatologist diagnostic knowledge into multimodal representations; DLC, which constrains the diagnostic reasoning during report generation; VSA, which establishes the correspondence between image features and imaging conclusions.}
  \vspace{-1.2em}
  \label{fig:overview}
\end{figure*}

As illustrated in Fig.~\ref{fig:overview}, NeoRed takes a neonatal 
CXR and clinical context as input, and generates a 
structured report comprising an imaging conclusion and a 
disease diagnosis.

Formally, the input CXR is denoted as $\mathbf{x}^{\text{cxr}} \in \mathbb{R}^{C \times H \times W}$ and the structured clinical context is denoted as $\mathbf{x}^{\text{txt}}$. Vision encoder $\mathcal{E}_v$ and a projector $\mathcal{P}$ extract and map visual features into a sequence of visual tokens $\mathbf{V} = \{v_1, v_2, \ldots, 
v_{N_v}\}$:
\begin{equation}
    \mathbf{V} = \mathcal{P}(\mathcal{E}_v(\mathbf{x}^{\text{cxr}})).
\end{equation}
The $\mathbf{x}^{\text{txt}}$ is tokenized into a sequence of textual tokens $\mathbf{T} = \{t_1, t_2, \ldots, t_{N_t}\}$. The two token sequences are concatenated into a joint multimodal input $\mathbf{U} = [\mathbf{V}, \mathbf{T}]$, which is fed into the LLM backbone to autoregressively generate target report $\mathcal{R} = \{r_1, r_2, \ldots, r_K\}$. K denotes the length of generated report in tokens. The model is optimized by minimizing the autoregressive cross-entropy loss over the ground-truth report $\mathcal{R}^* = \{r_1^*, r_2^*, \ldots, r_K^*\}$:
\begin{equation}
    P_{\Theta}(\mathcal{R} | \mathbf{U}) = \prod_{i=1}^{K} 
    P_{\Theta}(r_i | \mathbf{U}, r_{<i}),
    \label{eq:generation}
\end{equation}
\vspace{-0.6em}
\begin{equation}
    \mathcal{L}_\text{LM} = - \sum_{i=1}^{K} \log P_{\Theta}
    (r_i^* | \mathbf{U}, r_{<i}^*),
    \label{eq:loss_lm}
\end{equation}
where $r_{<i} = \{r_1, \ldots, r_{i-1}\}$ represents all previously generated tokens before step $i$, $\Theta$ denotes the learnable parameters.

\subsection{KLA: Knowledge-Logic-Alignment}
To emulate the diagnostic logic of experienced neonatologists and improve joint multimodal diagnosis, we design a novel KLA framework comprising KPI, DLC and VSA.

\noindent\textbf{1) Knowledge Prior Injection}

Neonatal respiratory disease diagnosis relies on both clinical context and CXRs, yet existing methods overlook the exploration of disease-specific modality dependencies and fail to effectively leverage this multidimensional information. To address this issue, we propose KPI to inject neonatologist-inspired priors into multimodal representations, implicitly guiding disease-specific attention of MLLM. 

Formally, for each modality $m \in \mathcal{M} = \{\text{dev, peri, phys, cxr}\}$, we first extract the input embeddings $\boldsymbol{h}^{m} \in \mathbb{R}^{B \times L \times D}$ from the corresponding developmental, perinatal, physiological, and image tokens.
\begin{table}[t]
\centering
\fontsize{8pt}{8pt}\selectfont
\renewcommand{\arraystretch}{0.66}
\setlength{\tabcolsep}{1.5pt}

\begin{tabular*}{0.98\columnwidth}{
@{\extracolsep{\fill}}lcccccccc@{}
}
\toprule
Modality & Pneu. & NRDS & TTN & PTx & BPD & Ate. & PE & Normal \\
\midrule
Dev.  & 0.2 & 1.0 & 0.5 & 0.3 & 1.0 & 0.3 & 0.4 & 1.0 \\
Peri. & 1.0 & 0.9 & 1.0 & 0.3 & 0.4 & 0.3 & 0.8 & 1.0 \\
Phys. & 1.0 & 1.0 & 0.9 & 1.0 & 0.8 & 1.0 & 1.0 & 1.0 \\
Img.  & 1.2 & 1.5 & 1.3 & 1.6 & 1.4 & 1.5 & 1.3 & 1.0 \\
\bottomrule
\end{tabular*}
\vspace{-0.8em}
\caption{Diagnostic priors between diseases and modalities. Disease abbreviations: Pneu. (Pneumonia), PTx (Pneumothorax), Ate. (Atelectasis), and PE (Pleural Effusion).}
\label{tab:priors}
\vspace{-1.6em}
\end{table}
Here, $L$ represents sequence length (i.e., number of tokens for each modality), and $D$ is embedding dimension. After applying average pooling along token dimension, we obtain modality-specific features $\bar{\boldsymbol{h}}^{m} \in \mathbb{R}^{B \times D}$. These features are then stacked to form a unified multimodal representation $\bar{\boldsymbol{H}} \in \mathbb{R}^{B \times |\mathcal{M}| \times D}$. To model dependencies between diseases and modalities, we introduce a learnable 
disease-specific prior matrix 
$\boldsymbol{W}_{prior}\in\mathbb{R}^{N\times |\mathcal{M}|}$, initialized using neonatologist-inspired priors (see Tab.~\ref{tab:priors}). The prior matrix $\boldsymbol{W}_{prior}$ is normalized via softmax and then used to weight modality-specific features $\bar{\boldsymbol{H}}$, yielding disease-specific multimodal representations $\boldsymbol{F}^{diag} \in \mathbb{R}^{B \times N \times D}$, where $N$ denotes the number of disease categories.
\begin{equation}
\boldsymbol{F}^{diag}
=
\operatorname{softmax}(\boldsymbol{W}_{prior}) \cdot \bar{\boldsymbol{H}},
\label{eq:diag}
\end{equation}
\begin{equation}
\boldsymbol{F}^{cli} = [\bar{\mathbf{h}}^{dev}; \bar{\mathbf{h}}^{peri}; \bar{\mathbf{h}}^{phys}].
\end{equation}
%


Then, $\mathbf{F}^{diag}$, $\bar{\mathbf{h}}^{cxr}$, and $\boldsymbol{F}^{cli}$ are passed through a dedicated classifier and supervised with binary cross-entropy (BCE) loss against disease labels, independently. This process yields prior classification loss $\mathcal{L}_{\text{prior}}$, CXR classification loss $\mathcal{L}_{\text{cxr}}$, and clinical classification loss $\mathcal{L}_{\text{cli}}$. The total optimization objective of KPI is formulated as:
\begin{equation}
\mathcal{L}_{\text{kpi}} = \mathcal{L}_{\text{prior}} + \lambda_1 \mathcal{L}_{\text{cxr}} + \lambda_2 \mathcal{L}_{\text{cli}},
\end{equation}
where $\lambda_1=\lambda_2=0.1$ are weighting coefficients.


\noindent\textbf{2) Diagnostic Logic Constraint}

Existing MLLMs generate reports in an autoregressive manner without explicit constraints ensuring that the generated text remains consistent with the underlying diagnostic decision. This may result in linguistically fluent yet diagnostically inconsistent reports. To address this issue, we propose DLC to ensure semantic consistency between generated text and clinical diagnostic logic.

To address this issue, we seek a global representation to regulate autoregressive generation. The BOS hidden state naturally serves this role in causal LLMs, as all subsequent tokens can access it through self-attention. Our attention analysis further shows strong token-level dependency on BOS, motivating us to inject diagnostic supervision into BOS and transform it into a global diagnostic anchor for guiding report generation. Specifically, we impose diagnostic supervision on BOS token and align its diagnostic distribution with multimodal diagnostic representations. We adopt multimodally fused features as local diagnostic representations, following the same formulation as $\boldsymbol{F}^{diag}$ in KPI (as defined in Eq.~\ref{eq:diag}), while features are derived from the last-layer hidden states of LLM. The hidden state of \texttt{<BOS>} token and $\boldsymbol{F}^{diag}$ are passed through dedicated classifiers and supervised with BCE loss, yielding global semantic loss $\mathcal{L}_{\text{gl}}$ and local classification loss $\mathcal{L}_{\text{lc}}$. To enforce semantic consistency between report generation and diagnostic logic, we compute Jensen-Shannon divergence between the two predictive distributions of global and local views as diagnostic consistency loss ${\mathcal{L}_{\text{dc}}}$.

\begin{table*}[t]
\centering
\renewcommand{\arraystretch}{0.8}
\setlength{\tabcolsep}{2.5pt}
\resizebox{\textwidth}{!}{
\begin{tabular}{llccccccccccc}
\toprule[1.5pt]
\multirow{2}{*}{\textbf{Type}}
& \multirow{2}{*}{\textbf{Model}}
& \multicolumn{6}{c}{\textbf{NLG}}
& \multicolumn{3}{c}{\textbf{CE}}
& \multirow{2}{*}{\textbf{Avg.}}
& \multirow{2}{*}{\textbf{$p$-value}} \\
\cmidrule(lr){3-8}
\cmidrule(lr){9-11}
&
& \textbf{ROUGE-L}
& \textbf{ROUGE-1}
& \textbf{BLEU-1}
& \textbf{BLEU-2}
& \textbf{METEOR}
& \textbf{RaTE}
& \textbf{F1}
& \textbf{P}
& \textbf{R}
& & \\
\midrule

\rowcolor{gray!5}
\multicolumn{13}{c}{\textit{NeoCXR}} \\
\midrule

\multirow{5}{*}{Generalist}
& LLaVA-NeXT-7B~\cite{Llava-next}
& 16.17 & 16.80 & 7.89 & 4.45 & 22.89
& 34.25 & 4.96 & 19.14 & 2.85 & 14.38 & 1.7E-502 \\

& InternVL-2.5-8B~\cite{Intern2.5-vl}
& 16.44 & 17.27 & 9.84 & 5.56 & 23.21
& 34.45 & 8.60 & 17.62 & 5.69 & 15.41 & 4.7E-468 \\

& Qwen2.5-VL-7B~\cite{Qwen2.5-vl}
& 9.59 & 10.04 & 5.00 & 2.79 & 17.35
& 35.01 & 15.03 & 20.67 & 11.81
& 14.14 & 2.0E-486 \\

& Qwen3-VL-8B~\cite{Qwen3-vl}
& 15.14 & 15.60 & 8.71 & 5.22 & 22.14
& 37.51 & 24.25 & 22.07
& 26.90 & 19.73 & 1.3E-387 \\

& Qwen3-VL-8B $^{\dagger}$ ~\cite{Qwen3-vl}
& \underline{52.28} & \underline{52.91} & \underline{48.27}
& \underline{42.38} & \underline{52.39} & \underline{58.39}
& \underline{62.32} & \underline{61.81} & \underline{62.83}
& \underline{54.85} & 6.1E-07 \\

\midrule

\multirow{4}{*}{Medical}
& LLaVA-Med-7B~\cite{Llava-med}
& 19.97 & 22.01 & 10.88 & 5.67 & 15.55
& 28.83 & 3.13 & 18.75 & 1.71 & 14.06 & 9.3E-513 \\

& LLaVA-Rad-7B~\cite{LLaVA-rad}
& 16.80 & 17.65 & 17.47 & 6.79 & 16.52
& 35.10 & 3.81 & 8.00 & 2.50 & 13.85 & 1.2E-481 \\

& HuatuoGPT-V-7B~\cite{Huatuogpt}
& 20.51 & 21.76 & 14.09
& 7.84 & 27.21
& 33.36 & 9.73 & 21.84 & 6.26
& 18.07 & 1.0E-451 \\

& Lingshu-7B~\cite{Lingshu}
& 16.80 & 17.92 & 9.22 & 5.06 & 23.47
& 33.33 & 6.29 & 16.13 & 3.91 & 14.68 & 2.9E-476 \\

& LLaVA-Rad-7B $^{\dagger}$~\cite{LLaVA-rad}
& 50.06 & 50.33 & 44.70 & 38.76 & 48.72
& 57.49 & 63.21 & 62.53 & 63.91 & 53.30 & 2.1E-09 \\

\midrule

\rowcolor{blue!5}
& \textbf{NeoRed}
& \textbf{53.29} & \textbf{53.56} & \textbf{48.83}
& \textbf{42.54} & \textbf{52.42} & \textbf{60.25}
& \textbf{65.19} & \textbf{64.62} & \textbf{65.77}
& \textbf{56.27} & — \\

\midrule[1.2pt]

\rowcolor{gray!5}
\multicolumn{13}{c}{\textit{NeoCXR-EV}} \\
\midrule
\multirow{5}{*}{Generalist}
& LLaVA-NeXT-7B~\cite{Llava-next}
& 14.85 & 15.63 & 8.37 & 4.71 & 22.05
& 29.91 & 6.79 & 45.05 & 3.67 & 16.78 & 4.3E-225 \\

& InternVL-2.5-8B~\cite{Intern2.5-vl}
& 16.58 & 17.53 & 11.98 & 7.06 & 25.08
& 31.08 & 14.43 & 38.54 & 8.88
& 19.02 & 5.3E-172 \\

& Qwen2.5-VL-7B~\cite{Qwen2.5-vl}
& 9.36 & 9.97 & 6.03 & 3.43 & 19.74
& 31.55 & 17.07 & 47.03 & 10.43
& 17.18 & 1.6E-205 \\

& Qwen3-VL-8B~\cite{Qwen3-vl}
& 15.51 & 16.35 & 10.50 & 6.24 & 23.77
& 34.63 & 34.32 & 42.18
& 28.93 & 23.60 & 1.5E-84 \\

& Qwen3-VL-8B $^{\dagger}$ ~\cite{Qwen3-vl}
& \underline{26.79} & \underline{27.12} & \underline{20.14}
& \underline{14.41} & \underline{25.78} & \underline{42.10}
& \underline{37.49} & 43.79 & \underline{32.77}
& \underline{30.05} & 3.7E-11 \\

\midrule

\multirow{4}{*}{Medical}
& LLaVA-Med-7B~\cite{Llava-med}
& 12.57 & 17.08 & 5.56 & 3.49 & 12.57
& 24.69 & 4.14 & \textbf{87.88} & 2.12
& 18.90 & 3.8E-266 \\

& LLaVA-Rad-7B~\cite{LLaVA-rad}
& 13.25 & 12.83 & 15.51 & 3.88 & 13.25
& 30.38 & 4.07 & 11.00 & 2.50
& 11.85 & 3.9E-238 \\

& HuatuoGPT-V-7B~\cite{Huatuogpt}
& 20.97 & 22.22 & 16.84
& 9.78 & \underline{28.94}
& 31.24 & 10.68 & 36.43 & 6.26
& 20.37 & 7.2E-161 \\

& Lingshu-7B~\cite{Lingshu}
& 16.64 & 17.61 & 10.76
& 6.20 & 23.96
& 30.52 & 8.57 & 41.94 & 4.77
& 17.89 & 1.1E-199 \\

& LLaVA-Rad-7B $^{\dagger}$ ~\cite{LLaVA-rad}
& 31.21 & 32.52 & 26.82
& 20.49 & 31.40 & 44.36
& 36.15 & 42.08 & 31.69
& 32.97 & 5.2E-1 \\

\midrule

\rowcolor{blue!5}
& \textbf{NeoRed}
& \textbf{32.90} & \textbf{33.13} & \textbf{27.12}
& \textbf{20.92} & \textbf{32.34} & \textbf{45.85}
& \textbf{39.28} & 44.36 & \textbf{35.24}
& \textbf{34.57} & — \\

\bottomrule[1.5pt]
\end{tabular}
}
\vspace{-0.8em}
\caption{Performance comparison between NeoRed and existing MLLMs on the NeoCXR and NeoCXR-EV datasets.
$\dagger$ is fine-tuned under the same settings of NeoRed. The best and second-best results are highlighted in \textbf{bold} and \underline{underline}.}
\vspace{-1.2em}
\label{tab:neocxr_all}
\end{table*}
\vspace{-0.5em}
\begin{equation}
\mathcal{L}_{\text{dc}}
=
\frac{1}{2}\mathrm{KL}(\boldsymbol{p}\|\boldsymbol{m})
+
\frac{1}{2}\mathrm{KL}(\boldsymbol{q}\|\boldsymbol{m}),
\end{equation}
where $\boldsymbol{p}$ and $\boldsymbol{q}$ denote the 
predicted probabilities from global and local views, and 
$\boldsymbol{m}$ denotes their mean. 

The optimization objective of DLC is defined as:
\begin{equation}
\mathcal{L}_{\text{dlc}}
=
\alpha\,\mathcal{L}_{\text{gl}}
+
\beta\,\mathcal{L}_{\text{lc}}
+
\gamma\,\mathcal{L}_{\text{dc}},
\end{equation}
where weighting coefficients $\alpha=1.0$, $\beta=0.5$, $\gamma=0.1$.

\begin{table}[t]
\centering
\scriptsize
\setlength{\tabcolsep}{2.0pt}
\renewcommand{\arraystretch}{0.66}
\resizebox{\columnwidth}{!}{
\begin{tabular}{lccccccc}
\toprule
Model & ROUGE-L & METEOR & RaTE & F1 & P & R & Avg. \\
\midrule
Baseline & 50.06 & 48.72 & 57.49 & 63.21 & 62.53 & 63.91 & 57.65 \\
w/o KPI  & 52.33 & 50.73 & 58.74 & 64.18 & 63.53 & 64.84 & 59.06 \\
w/o DLC  & 52.67 & 51.79 & 59.09 & 63.54 & 62.90 & 64.20 & 59.03 \\
w/o VSA  & 51.60 & 51.32 & 59.29 & 65.14 & 64.53 & 65.77 & 59.61 \\
\midrule
\rowcolor{blue!5}
\textbf{NeoRed}
& \textbf{53.29}
& \textbf{52.42}
& \textbf{60.25}
& \textbf{65.19}
& \textbf{64.62}
& \textbf{65.77}
& \textbf{60.26} \\
\bottomrule
\end{tabular}
}
\vspace{-0.8em}
\caption{Ablation study of KLA on the NeoCXR dataset.}
\vspace{-1em}
\label{tab:ablation_neocxr}
\end{table}

\begin{table}[t]
\centering
\scriptsize
\setlength{\tabcolsep}{2.0pt}
\renewcommand{\arraystretch}{0.66}
\resizebox{\columnwidth}{!}{
\begin{tabular}{lccccccc}
\toprule
Model & ROUGE-L & METEOR & RaTE & F1 & P & R & Avg. \\
\midrule
Baseline
& 50.06 & 48.72 & 57.49 & 63.21 & 62.53 & 63.91 & 57.65 \\

w/o $\mathcal{L}_{\text{prior}}$
& 52.45 & 51.53 & 59.00 & 64.45 & 63.90 & 65.01 & 59.39 \\

w/o $\mathcal{L}_{\text{cli}}$
& 52.84 & 51.89 & 59.05 & 64.24 & 63.65 & 64.84 & 59.42 \\

w/o $\mathcal{L}_{\text{cxr}}$
& 53.04 & 52.22 & 59.16 & 64.75 & 64.53 & 64.98 & 59.78 \\

\midrule
\rowcolor{blue!5}
\textbf{NeoRed}
& \textbf{53.29}
& \textbf{52.42}
& \textbf{60.25}
& \textbf{65.19}
& \textbf{64.62}
& \textbf{65.77}
& \textbf{60.26} \\
\bottomrule
\end{tabular}
}
\vspace{-0.8em}
\caption{Internal ablation of KPI on the NeoCXR dataset.}
\vspace{-2.2em}
\label{tab:ablation_neocxr_prior}
\end{table}

\noindent\textbf{3) Visual Semantic Alignment}

To further align visual evidence with its clinical textual interpretation, we design VSA, which establishes bidirectional semantic correspondence between image features and imaging conclusions, encouraging the generated imaging conclusions to be supported by visual evidence.


We extract the last-layer hidden states of image tokens by the LLM, 
denoted as $\boldsymbol{h}^{cxr}\in\mathbb{R}^{B\times L\times D}$, 
and apply average pooling along the token dimension to obtain image 
representation $\bar{\boldsymbol{h}}^{cxr}$.
We extract the textual representation of imaging conclusion from the ground-truth report 
$R^* = \{r^*_1, r^*_2, \dots, r^*_K\}$ by applying a binary mask over token-level hidden states, 
where tokens in the imaging conclusion field are set to 1 and others to 0. Textual representation $\bar{\boldsymbol{h}}^{con}$ is obtained via average pooling over masked tokens. We employ a bidirectional contrastive objective to align image and text representations:
\begin{table}[t]
\centering
\scriptsize
\setlength{\tabcolsep}{2.0pt}
\renewcommand{\arraystretch}{0.66}
\resizebox{\columnwidth}{!}{
\begin{tabular}{lccccccc}
\toprule
Model & ROUGE-L & METEOR & RaTE & F1 & P & R & Avg. \\
\midrule
Baseline
& 50.06 & 48.72 & 57.49 & 63.21 & 62.53 & 63.91 & 57.65 \\

w/o $\mathcal{L}_{\text{gl}}$
& 52.06 & 52.11 & 59.47 & 64.14 & 63.51 & 64.78 & 59.34 \\

w/o $\mathcal{L}_{\text{lc}}$
& 52.25 & 52.34 & 59.56 & 64.11 & 63.40 & 64.84 & 59.42 \\

w/o $\mathcal{L}_{\text{dc}}$
& 52.31 & 52.41 & 60.09 & 64.51 & 64.56 & 64.47 & 59.74 \\

\midrule
\rowcolor{blue!5}
\textbf{NeoRed}
& \textbf{53.29}
& \textbf{52.42}
& \textbf{60.25}
& \textbf{65.19}
& \textbf{64.62}
& \textbf{65.77}
& \textbf{60.26} \\
\bottomrule
\end{tabular}
}
\vspace{-0.8em}
\caption{Internal ablation of DLC on the NeoCXR dataset.}
\vspace{-1em}
\label{tab:ablation_neocxr_diag}
\end{table}
\begin{figure}[h]
  \centering
\includegraphics[width=1.0\linewidth]{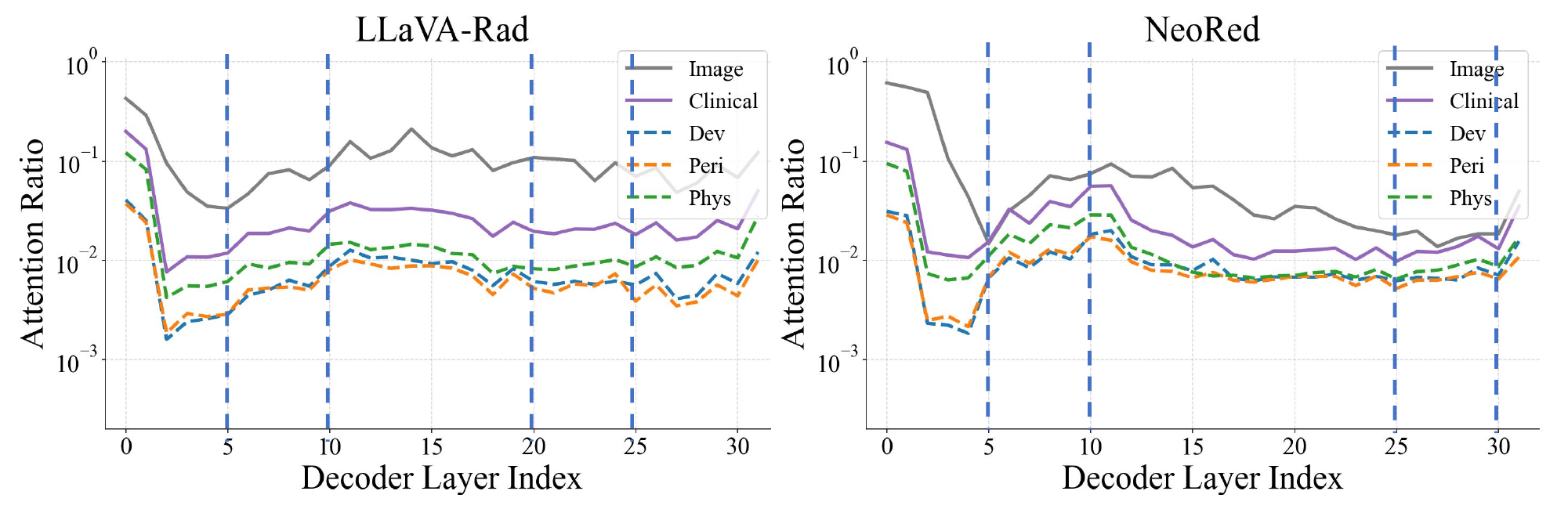}
  \vspace{-1.8em}
  \caption{Attention Analysis of generated tokens.}
  \label{fig:attention}
  \vspace{-1.5em}
\end{figure}
\begin{equation}
\mathcal{L}_{\mathrm{i2t}}
=
-\frac{1}{B}\sum_{i=1}^{B}
\log
\frac{\exp(s_{ii}/\tau)}
{\sum_{j=1}^{B}\exp(s_{ij}/\tau)},
\end{equation}

\begin{equation}
\mathcal{L}_{\mathrm{t2i}}
=
-\frac{1}{B}\sum_{i=1}^{B}
\log
\frac{\exp(s_{ii}/\tau)}
{\sum_{j=1}^{B}\exp(s_{ji}/\tau)},
\end{equation}

\begin{equation}
\mathcal{L}_{\mathrm{vsa}}
=
\frac{1}{2}
\left(
\mathcal{L}_{\mathrm{i2t}}
+
\mathcal{L}_{\mathrm{t2i}}
\right),
\end{equation}
where $s_{ij}$ denotes the cosine similarity, $\tau$ is set to $0.07$.


\begin{table*}[h]
\centering
\renewcommand{\arraystretch}{0.8}
\resizebox{\textwidth}{!}{
\begin{tabular}{llcccccccc}
\toprule[1.5pt]
\textbf{Type}
& \textbf{Model}
& \textbf{ROUGE-L}
& \textbf{ROUGE-1}
& \textbf{BLEU-1}
& \textbf{BLEU-2}
& \textbf{METEOR}
& \textbf{RaTE}
& \textbf{Avg.}
& \textbf{$p$-value} \\
\midrule
















\rowcolor{gray!5}
\multicolumn{10}{c}{\textit{MIMIC-CXR}} \\
\midrule

\multirow{5}{*}{Generalist}

& LLaVA-NeXT-7B~\cite{Llava-next}
& 17.14 & 18.15 & 10.51 & 3.72 & 14.99 & 41.49 & 17.67 & 4.5E-406 \\

& InternVL-2.5-8B~\cite{Intern2.5-vl}
& 23.15 & 24.53 & 21.53 & 8.70 & 20.30 & 45.89 & 24.02 & 9.7E-59 \\

& Qwen2.5-VL-7B~\cite{Qwen2.5-vl}
& 24.33 & 25.90 & 22.30 & 9.26
& 19.28 & 45.74 & 24.47 & 4.0E-44 \\

& Qwen3-VL-8B~\cite{Qwen3-vl}
& 23.31 & 24.80 & 20.96 & 8.70
& \textbf{21.16} & 48.33 & 24.54 & 8.0E-40 \\


\midrule

\multirow{4}{*}{Medical}
& LLaVA-Med-7B~\cite{Llava-med}
& 14.39 & 14.91 & 3.70 & 0.68 & 6.07 & 32.67 & 12.07 & 6.2E-889 \\

& LLaVA-Rad-7B~\cite{LLaVA-rad}
& 28.94 & \underline{30.45} & \textbf{24.06}
& \textbf{12.88} & \underline{21.05} & \textbf{53.63}
& \textbf{28.50} & 1.3E-20 \\

& HuatuoGPT-V-7B~\cite{Huatuogpt}
& 23.11 & 24.56 & 21.35 & 8.85
& 19.13 & 47.96 & 24.16 & 1.9E-53 \\

& Lingshu-7B~\cite{Lingshu}
& \textbf{29.71} & \textbf{30.89} & 20.58
& 9.75 & 18.53 & \underline{50.76}
& 26.70 & 8.3E-1 \\


\midrule

\rowcolor{blue!5}
& \textbf{NeoRed}
& \underline{29.01} & 29.80 & \underline{22.49}
& \underline{11.37} & 20.30 & 47.43
& \underline{26.73} & —\\

\midrule[1.2pt]

\rowcolor{gray!5}
\multicolumn{10}{c}{\textit{IU-Xray}} \\
\midrule
\multirow{5}{*}{Generalist}
& LLaVA-1.5-7B~\cite{llava-1.5}
& 13.95 & 15.26 & 13.94 & 4.37 & 16.23 & 40.07 & 17.30 & 2.7E-860 \\

& LLaVA-NeXT-7B~\cite{Llava-next}
& 16.17 & 16.80 & 7.89 & 4.45 & 22.89 & 34.25 & 17.08 & 3.7E-224 \\

& InternVL-2.5-8B~\cite{Intern2.5-vl}
& 24.78 & 26.51 & 21.83 & 9.18 & 24.25 & 51.63 & 26.36 & 6.5E-209 \\

& Qwen2.5-VL-7B~\cite{Qwen2.5-vl}
& 32.62 & 34.46 & 27.64 & 14.31
& 27.59 & 54.62 & 31.87 & 3.8E-92 \\

& Qwen3-VL-8B~\cite{Qwen3-vl}
& 27.62 & 29.38 & 24.53 & 11.73
& \textbf{28.07} & 51.78 & 28.85 & 2.7E-27 \\


\midrule

\multirow{4}{*}{Medical}
& LLaVA-Med-7B~\cite{Llava-med}
& 18.21 & 18.51 & 7.66 & 2.18 & 8.05 & 35.12 & 14.96 & 3.7E-1104 \\

& LLaVA-Rad-7B~\cite{LLaVA-rad}
& 33.11 & 35.81 & 27.46
& 15.18 & 23.33 & \textbf{61.14}
& 32.67 & 3.1E-29 \\

& HuatuoGPT-V-7B~\cite{Huatuogpt}
& 27.05 & 28.34 & 22.10 & 10.52
& 25.84 & 53.30 & 27.86 & 4.7E-32 \\

& Lingshu-7B~\cite{Lingshu}
& \textbf{41.33} & \textbf{43.51} & \textbf{30.42}
& \textbf{20.15} & \underline{27.64} & \underline{57.32}
& \textbf{36.73} & 7.1E-861 \\













\midrule

\rowcolor{blue!5}
& \textbf{NeoRed}
& \underline{36.27} & \underline{37.75} & \underline{28.27}
& \underline{15.43} & 24.03 & 56.33
& \underline{33.01} & — \\

\bottomrule[1.5pt]
\end{tabular}
}
\vspace{-0.8em}
\caption{Performance comparison between NeoRed and existing MLLMs on the MIMIC-CXR and IU-Xray datasets.}

\vspace{-1.2em}
\label{tab:adult_datasets}
\end{table*}

\subsection{Training Objective}
The total training objective $\mathcal{L}_\text{total}$ is formulated as a weighted sum of language modeling loss and three auxiliary losses:
\begin{equation}
    \mathcal{L}_\text{total} = w_\text{lm}\,\mathcal{L}_\text{LM} + w_\text{kpi}\,\mathcal{L}_\text{kpi} + w_\text{dlc}\,\mathcal{L}_\text{dlc} + w_\text{vsa}\,\mathcal{L}_\text{vsa},
\end{equation}
where $w_\text{lm}=1.0$ and $w_\text{kpi}$ = $w_\text{dlc}$ = $w_\text{vsa}$ = $0.5$.


\begin{table}[h]
\centering
\scriptsize
\setlength{\tabcolsep}{2pt}
\renewcommand{\arraystretch}{0.66}
\resizebox{\columnwidth}{!}{
\begin{tabular}{lccccccc}
\toprule
Type
& ROUGE-L
& METEOR
& RaTE
& F1
& P
& R
& Avg. \\
\midrule
Random
& 50.81
& 48.67
& 58.15
& 61.18
& 60.53
& 61.84
& 56.86 \\

All-one
& 51.94
& 49.94
& 58.67
& 61.28
& 59.80
& 62.84
& 57.41 \\

All-zero
& 50.26
& 48.10
& 57.14
& 60.49
& 58.95
& 62.12
& 56.18 \\
\midrule
\rowcolor{blue!5}
\textbf{Priors}
& \textbf{53.29}
& \textbf{52.42}
& \textbf{60.25}
& \textbf{65.19}
& \textbf{64.62}
& \textbf{65.77}
& \textbf{60.26} \\
\bottomrule
\end{tabular}
}
\vspace{-0.8em}
\caption{Ablation of prior matrix on NeoCXR.}
\vspace{-0.5em}
\label{tab:prior}
\end{table}
\begin{table}[t]
\centering
\scriptsize
\setlength{\tabcolsep}{3.2pt}
\renewcommand{\arraystretch}{0.66}
\resizebox{\columnwidth}{!}{
\begin{tabular}{lccccccc}
\toprule
Weight
& ROUGE-L
& METEOR
& RaTE
& F1
& P
& R
& Avg. \\
\midrule
0.25
& 52.15
& 50.47
& 58.79
& 63.54
& 63.88
& 63.20
& 58.67 \\

\midrule
\rowcolor{blue!5}
\textbf{0.50}
& \textbf{53.29}
& \textbf{52.42}
& \textbf{60.25}
& \textbf{65.19}
& \textbf{64.62}
& \textbf{65.77}
& \textbf{60.26} \\

\midrule
0.75
& 50.04
& 47.77
& 57.98
& 63.40
& 63.18
& 63.63
& 57.67 \\
\bottomrule
\end{tabular}
}
\vspace{-0.8em}
\caption{Sensitivity analysis of loss weights.}
\label{tab:loss}
\vspace{-1.5em}
\end{table}


\section{Experiments}
\subsection{Datasets and Evaluation Metrics}
We evaluate our model on two benchmarks. 1) The neonatal benchmark
comprises internal test set of NeoCXR (1434 samples) and the full 
NeoCXR-EV (1089 samples), as we mentioned in Tab.~\ref{tab:datasets}. 2) The adult benchmarksinclude the official test sets of MIMIC-CXR~\cite{MIMIC-CXR} and IU-Xray~\cite{IU-Xray}. After removing samples with empty findings or impression sections, 2,737 and 3,193 samples remain for MIMIC-CXR and IU-Xray. We evaluate generated reports using natural language generation (NLG) metrics, including ROUGE-L \cite{RougeL}, BLEU-1~\cite{Bleu}, METEOR~\cite{METEOR}, RaTE~\cite{Ratescore}, to assess linguistic quality and medical factuality. For neonatal benchmarks, we evaluate diagnostic consistency using an unified label extraction protocol, reporting micro precision, recall, and F1 scores. Details and validation are provided in \textit{supplementary material}. For adult benchmarks, we follow previous report generation studies~\cite{Lingshu,beyond} and report standard NLG together with RaTE for factuality evaluation. Implementation details provided in \textit{supplementary material}.

\section{Results}
\subsection{Compare with SOTA MLLMs}
We compare NeoRed against 8 state-of-the-art (SOTA) MLLMs across two domains. 1) Generalist models:  LLaVA-NeXT-7B~\cite{Llava-next}, Qwen2.5-VL-7B~\cite{Qwen2.5-vl}, InternVL-2.5-8B~\cite{Intern2.5-vl}, and Qwen3-VL-8B~\cite{Qwen3-vl}. 2) Medical models: LLaVA-Med-7B~\cite{Llava-med}, LLaVA-Rad-7B~\cite{LLaVA-rad}, HuatuoGPT-V-7B~\cite{Huatuogpt}, Lingshu-7B~\cite{Lingshu}.
\begin{figure}[!t]
  \centering
  \includegraphics[width=1.0\linewidth]{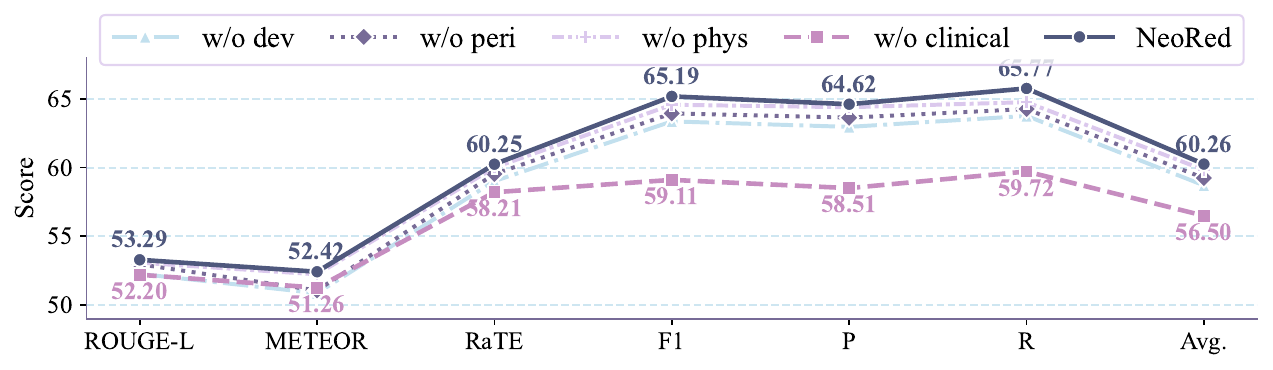}
  \vspace{-1.6em}
  \caption{Ablation of clinical context.}
  \label{fig:clinical}
  \vspace{-1.5em}
\end{figure}
Following the common practice in prior works \cite{Lingshu,medmo,eyecaregpt}, we evaluate existing models in a zero-shot setting by directly using their publicly released weights under a unified prompt, without any additional fine-tuning. Detailed prompt templates are provided in \textit{supplementary material}. 


\textbf{1) Performance on neonatal benchmark.}
Tab.~\ref{tab:neocxr_all} reports results on the in-domain NeoCXR test set and the external NeoCXR-EV set. On NeoCXR, NeoRed substantially outperforms both the best generalist model, Qwen3-VL-8B (24.25\% F1), and the strongest medical model, HuatuoGPT-V-7B (9.73\% F1). NeoRed also remains the best-performing model on NeoCXR-EV, demonstrating generalization under substantial disease distribution shift (see supplementary material). All pair-wise comparisons of the average performance achieve statistical significance ($p<0.05$).

To demonstrate the contribution of neonatal-domain adaptation, we fine-tune Qwen3-VL and LLaVA-Rad under the same training settings as NeoRed. Fine-tuning on NeoCXR substantially enhances their performance, achieving average gains of 35.12\% and 39.45\% on NeoCXR, and 6.45\% and 21.12\% on NeoCXR-EV for Qwen3-VL-8B and LLaVA-Rad-7B, respectively, highlighting the importance of neonatal-specific data adaptation. Furthermore, equipped with the proposed KLA framework, NeoRed consistently outperforms fine-tuned LLaVA-Rad on NeoCXR by 2.97\% and Qwen3-VL on NeoCXR-EV by 4.52\%.


 
\textbf{2) Generalization to adult benchmark.}
Tab.~\ref{tab:adult_datasets} evaluates NeoRed on adult benchmarksMIMIC-CXR and IU-Xray. On MIMIC-CXR, NeoRed achieves an average score of 26.73\%, second only to LLaVA-Rad-7B, which is specifically optimized for adult chest X-rays. On IU-Xray, NeoRed achieves an average score of 33.01\%, second only to the best model, Lingshu. These results indicate that NeoRed retains generalization on adult report generation.

\begin{figure}[!t]
  \centering
  \includegraphics[width=1.0\linewidth]{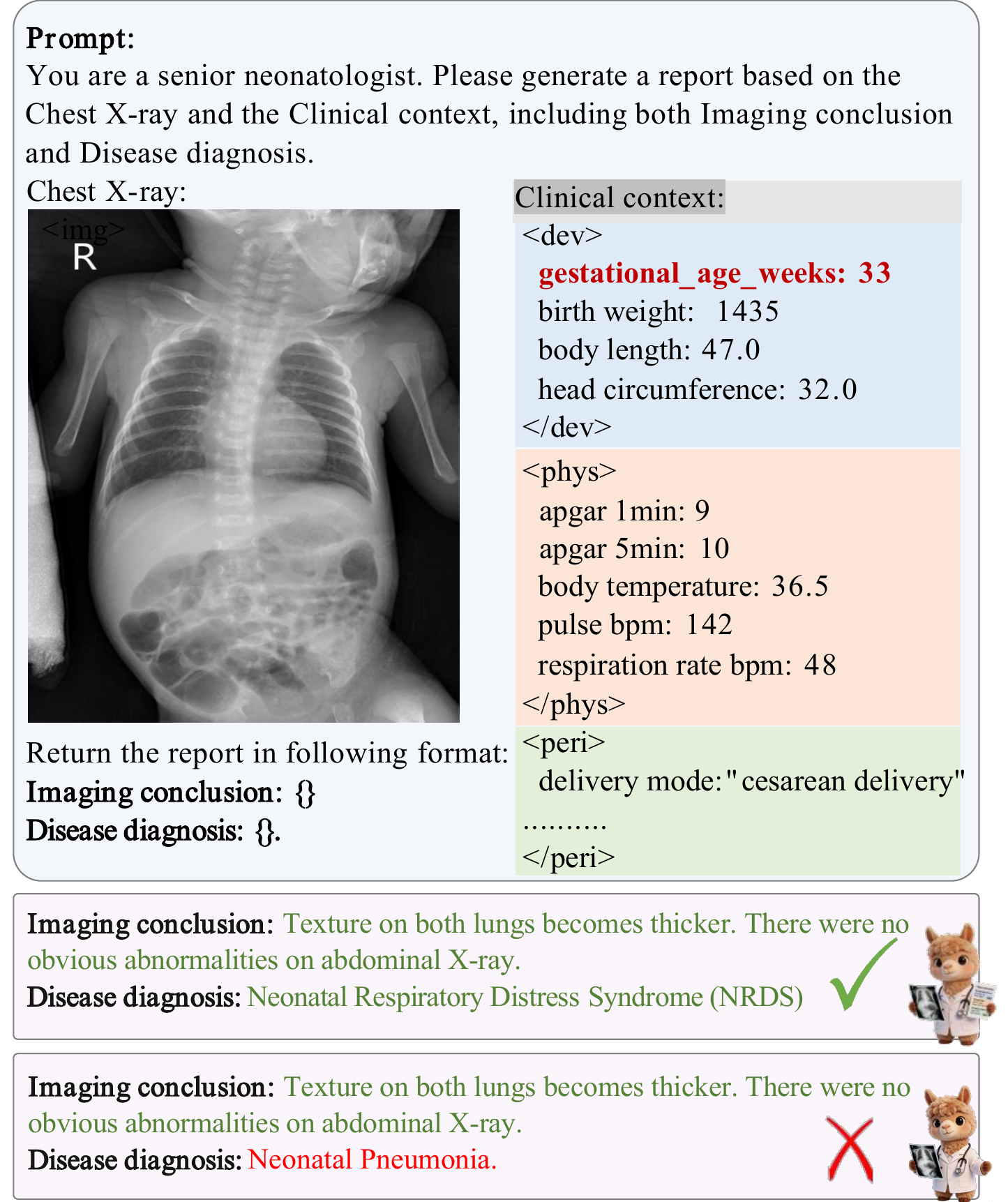}
  \vspace{-1.5em}
  \caption{Diagnostic case w/ and w/o clinical context.}
  \label{fig:case}
  \vspace{-1.8em}
\end{figure}


See comparative case studies  in \textit{supplementary material}.

\subsection{Ablation Studies}
To comprehensively evaluate the effectiveness of each module in the proposed KLA framework, we conduct extensive ablation studies on NeoCXR dataset as follows: 


\textbf{1) Ablation of KLA framework.} Our baseline is LLaVA-Rad-7B fine-tuned on NeoCXR without KLA. We adopt a leave-one-out ablation strategy to isolate each module’s contribution, while comparison with baseline demonstrates their joint effectiveness. As shown in Tab.~\ref{tab:ablation_neocxr}, removing KPI or DLC module causes a larger drop in CE metrics, confirming the importance of knowledge prior injection and diagnostic consistency constraint in neonatal disease diagnosis. Removing VSA primarily hurts NLG performance, confirming significance of vision alignment.

\textbf{2) Internal Ablation of KPI and DLC.}
Tab.~\ref{tab:ablation_neocxr_prior} evaluates the three KPI components. Removing $\mathcal{L}_\text{prior}$ degrades both NLG and CE performance, confirming the importance of prior knowledge injection. Excluding $\mathcal{L}*\text{cli}$ or $\mathcal{L}*\text{cxr}$ also degrades performance, confirming the benefit of modality-specific supervision. Tab.~\ref{tab:ablation_neocxr_diag} analyzes the three DLC components. Removing $\mathcal{L}_\text{gl}$ consistently harms both NLG and CE metrics, while excluding $\mathcal{L}_\text{dc}$ or $\mathcal{L}_\text{lc}$ also causes performance drops, with $\mathcal{L}_\text{lc}$ having the largest overall impact.


\textbf{3) Ablation of priors.}
The prior matrix is constructed based on disease–modality relevance determined through expert consensus between two neonatologists, with any disagreements resolved through discussion until consensus is reached. Comparison results of random, all-one, all-zero initializations confirm effectiveness of our priors (see Tab.~\ref{tab:prior}).

\textbf{4) Attention analysis of generated tokens.}
Following~\cite{sample100}, we sample 100 NeoCXR cases and compute the attention ratios of generated tokens to image and clinical tokens across decoder layers, averaged over heads and tokens. During the learning (layers 5–10) and diagnostic stages (layers 25–31), NeoRed achieves more balanced multimodal fusion than the vision-dominated LLaVA-Rad (see Fig.~\ref{fig:attention}), providing visual evidence for the effectiveness of our proposed KLA framework. Detailed token-level analysis is provided in \textit{supplementary material}.






\textbf{5) Sensitivity analysis of loss weight.}
Following prior work \cite{loss}, we adopt commonly used auxiliary-loss weights and validate them through sensitivity analysis. A representative analysis for weighting coefficient in overall objective ($w_\text{kpi}=w_\text{dlc}=w_\text{vsa}$=0.5) is shown in Tab.~\ref{tab:loss}.


\textbf{6) Ablation of clinical context.}
As shown in Fig.~\ref{fig:clinical}, removing all clinical context produces the worst results, demonstrating the necessity of clinical context. Removing any clinical category degrades performance, confirming contribution of each. Developmental factors have the greatest impact on CE metrics, followed by perinatal risks. A diagnostic case with and without clinical text is shown in Fig.~\ref{fig:case}. Image-only input leads to a pneumonia misdiagnosis, while incorporating clinical context enables correct NRDS diagnosis.

\section{Conclusion}
\label{c5}
We construct two real-world neonatal datasets (NeoCXR and NeoCXR-EV) and propose NeoRed, to the best of our knowledge, the first MLLM tailored for neonatal respiratory disease diagnosis, filling a critical gap in this domain. NeoRed organizes key clinical indicators into three categories (developmental factors, perinatal risks, and physiological status), providing structured clinical context for diagnosis. To further enhance multimodal diagnosis, we propose a novel KLA framework with three modules: KPI for injecting prior knowledge into multimodal representations, DLC for enforcing diagnostic consistency during report generation, and VSA for aligning visual features with textual descriptions. KLA incorporates neonatologist-inspired diagnostic priors and consistency constraints, forming a closed loop from knowledge acquisition to report generation. Extensive experiments show that NeoRed consistently outperforms mainstream MLLMs on neonatal benchmarks while maintaining  generalization on adult benchmarks (MIMIC-CXR and IU-Xray).

\bibliography{aaai2027}

@String{Computer = "{IEEE} Computer" }

@String{Academic = "Academic Press" }

@article{Llava-med,
  title={Llava-med: Training a large language-and-vision assistant for biomedicine in one day},
  author={Li, Chunyuan and Wong, Cliff and Zhang, Sheng and Usuyama, Naoto and Liu, Haotian and Yang, Jianwei and Naumann, Tristan and Poon, Hoifung and Gao, Jianfeng},
  journal={Advances in Neural Information Processing Systems},
  volume={36},
  pages={28541--28564},
  year={2023}
}

@inproceedings{llava-1.5,
  title={Improved baselines with visual instruction tuning},
  author={Liu, Haotian and Li, Chunyuan and Li, Yuheng and Lee, Yong Jae},
  booktitle={Proceedings of the IEEE/CVF conference on computer vision and pattern recognition},
  pages={26296--26306},
  year={2024}
}

@article{Neo1,
  title={Deep-learning-based multi-class classification for neonatal respiratory diseases on chest radiographs in neonatal intensive care units},
  author={Cho, Hye Won and Jung, Sumin and Park, Kyu Hee and Choi, Jin Wha and Heo, Ju Sun and Kim, Jaeyoung and Yun, Heerim and Yu, Donghoon and Son, Jinho and Choi, Byung Min},
  journal={Neonatology},
  volume={122},
  number={4},
  pages={446--454},
  year={2025}
}

@article{Neo2,
  title={Clinical characteristics and outcomes in neonates with perinatal acute respiratory distress syndrome in China: a national, multicentre, cross-sectional study},
  author={Chen, Long and Li, Jie and Shi, Yuan},
  journal={EClinicalMedicine},
  volume={55},
  year={2023},
  publisher={Elsevier}
}

@article{Umit,
  title={Umit: Unifying medical imaging tasks via vision-language models},
  author={Yu, Haiyang and Yi, Siyang and Niu, Ke and Zhuo, Minghan and Li, Bin},
  journal={arXiv preprint arXiv:2503.15892},
  year={2025}
}

@article{LLaVA-rad,
  title={Towards a clinically accessible radiology foundation model: open-access and lightweight, with automated evaluation},
  author={Chaves, Juan Manuel Zambrano and Huang, Shih-Cheng and Xu, Yanbo and Xu, Hanwen and Usuyama, Naoto and Zhang, Sheng and Wang, Fei and Xie, Yujia and Khademi, Mahmoud and Yang, Ziyi and others},
  journal={arXiv preprint arXiv:2403.08002},
  year={2024}
}

@inproceedings{llava-ultra,
  title={Llava-ultra: Large chinese language and vision assistant for ultrasound},
  author={Guo, Xuechen and Chai, Wenhao and Li, Shi-Yan and Wang, Gaoang},
  booktitle={Proceedings of the 32nd ACM international conference on multimedia},
  pages={8845--8854},
  year={2024}
}

@inproceedings{sample100,
  title={Boosting multimodal large language models with visual tokens withdrawal for rapid inference},
  author={Lin, Zhihang and Lin, Mingbao and Lin, Luxi and Ji, Rongrong},
  booktitle={Proceedings of the AAAI Conference on Artificial Intelligence},
  volume={39},
  number={5},
  pages={5334--5342},
  year={2025}
}

@article{Lingshu,
  title={Lingshu: A Generalist Foundation Model for Unified Multimodal Medical Understanding and Reasoning},
  author={Xu, Weiwen and Chan, Hou Pong and Li, Long and Aljunied, Mahani and Yuan, Ruifeng and Wang, Jianyu and Xiao, Chenghao and Chen, Guizhen and Liu, Chaoqun and Li, Zhaodonghui and others},
  journal={arXiv preprint arXiv:2506.07044},
  year={2025}
}

@article{Huatuogpt,
  title={Huatuogpt-vision, towards injecting medical visual knowledge into multimodal llms at scale},
  author={Chen, Junying and Gui, Chi and Ouyang, Ruyi and Gao, Anningzhe and Chen, Shunian and Chen, Guiming Hardy and Wang, Xidong and Zhang, Ruifei and Cai, Zhenyang and Ji, Ke and others},
  journal={arXiv preprint arXiv:2406.19280},
  year={2024}
}

@inproceedings{eyecaregpt,
  title={Eyecaregpt: Boosting comprehensive ophthalmology understanding with tailored dataset, benchmark and model},
  author={Li, Sijing and Lin, Tianwei and Lin, Lingshuai and Zhang, Wenqiao and Liu, Jiang and Yang, Xiaoda and Li, Juncheng and He, Yucheng and Song, Xiaohui and Xiao, Jun and others},
  booktitle={Proceedings of the 33rd ACM International Conference on Multimedia},
  pages={3893--3902},
  year={2025}
}

@article{Llava-next,
  title={Llava-next: Improved reasoning, ocr, and world knowledge, January 2024},
  author={Liu, Haotian and Li, Chunyuan and Li, Yuheng and Li, Bo and Zhang, Yuanhan and Shen, Sheng and Lee, Yong Jae},
  volume={1},
  number={8},
  year={2024}
}

@article{medmo,
  title={MedMO: Grounding and Understanding Multimodal Large Language Model for Medical Images},
  author={Deria, Ankan and Kumar, Komal and Dukre, Adinath Madhavrao and Segal, Eran and Khan, Salman and Razzak, Imran},
  journal={arXiv preprint arXiv:2602.06965},
  year={2026}
}

@article{bai2023qwen,
  title={Qwen technical report},
  author={Bai, Jinze and Bai, Shuai and Chu, Yunfei and Cui, Zeyu and Dang, Kai and Deng, Xiaodong and Fan, Yang and Ge, Wenbin and Han, Yu and Huang, Fei and others},
  journal={arXiv preprint arXiv:2309.16609},
  year={2023}
}

@inproceedings{medplib,
  title={Towards a multimodal large language model with pixel-level insight for biomedicine},
  author={Huang, Xiaoshuang and Shen, Lingdong and Liu, Jia and Shang, Fangxin and Li, Hongxiang and Huang, Haifeng and Yang, Yehui},
  booktitle={Proceedings of the AAAI Conference on Artificial Intelligence},
  volume={39},
  number={4},
  pages={3779--3787},
  year={2025}
}

@inproceedings{medclip,
  title={Medclip: Contrastive learning from unpaired medical images and text},
  author={Wang, Zifeng and Wu, Zhenbang and Agarwal, Dinesh and Sun, Jimeng},
  booktitle={Proceedings of the 2022 Conference on Empirical Methods in Natural Language Processing},
  pages={3876--3887},
  year={2022}
}

@inproceedings{medvqa2,
  title={Lapa: Latent prompt assist model for medical visual question answering},
  author={Gu, Tiancheng and Yang, Kaicheng and Liu, Dongnan and Cai, Weidong},
  booktitle={Proceedings of the IEEE/CVF Conference on Computer Vision and Pattern Recognition},
  pages={4971--4980},
  year={2024}
}

@inproceedings{blip,
  title={Blip: Bootstrapping language-image pre-training for unified vision-language understanding and generation},
  author={Li, Junnan and Li, Dongxu and Xiong, Caiming and Hoi, Steven},
  booktitle={International conference on machine learning},
  pages={12888--12900},
  year={2022},
  organization={PMLR}
}

@ARTICLE{Dialog,
  author={Sun, Yuanyuan and Zhou, Ting},
  journal={IEEE Access}, 
  title={DialogueMLLM: Transforming Multimodal Emotion Recognition in Conversation Through Instruction-Tuned MLLM}, 
  year={2025},
  volume={13},
  number={},
  pages={121048-121060}}

@article{Dialogshikra,
  title={Shikra: Unleashing multimodal llm's referential dialogue magic},
  author={Chen, Keqin and Zhang, Zhao and Zeng, Weili and Zhang, Richong and Zhu, Feng and Zhao, Rui},
  journal={arXiv preprint arXiv:2306.15195},
  year={2023}
}

@article{pefomed,
  title={PeFoMed: Parameter efficient fine-tuning of multimodal large language models for medical imaging},
  author={He, Jinlong and Li, Pengfei and Liu, Gang and He, Genrong and Chen, Zhaolin and Zhong, Shenjun},
  journal={arXiv preprint arXiv:2401.02797},
  year={2024}
}

@article{vlmrg,
  title={Maira-1: A specialised large multimodal model for radiology report generation},
  author={Hyland, Stephanie L and Bannur, Shruthi and Bouzid, Kenza and Castro, Daniel C and Ranjit, Mercy and Schwaighofer, Anton and P{\'e}rez-Garc{\'\i}a, Fernando and Salvatelli, Valentina and Srivastav, Shaury and Thieme, Anja and others},
  journal={arXiv preprint arXiv:2311.13668},
  year={2023}
}

@article{instructblip,
  title={Instructblip: Towards general-purpose vision-language models with instruction tuning},
  author={Dai, Wenliang and Li, Junnan and Li, Dongxu and Tiong, Anthony and Zhao, Junqi and Wang, Weisheng and Li, Boyang and Fung, Pascale N and Hoi, Steven},
  journal={Advances in neural information processing systems},
  volume={36},
  pages={49250--49267},
  year={2023}
}

@article{neocxr,
  title={Lung Ultrasound versus Chest X-ray for Diagnosing Pulmonary Disorders in Neonatal Age Group},
  author={Ismaiel, Hadeel Samir and Farouk, Hesham Mohamed and Mohammed, Mariam Hamdy},
  journal={Benha Medical Journal},
  volume={42},
  number={8},
  pages={80--92},
  year={2025},
  publisher={Benha University, Faculty of Medicine}
}

@article{neo_nomogram,
  title={A nomogram for predicting neonatal acute respiratory distress syndrome in patients with neonatal pneumonia after 34 weeks of gestation},
  author={Yu, Aosong and Hou, Huanhuan and Ran, Lingyi and Sun, Xiaojia and Xin, Wanchun and Feng, Tong},
  journal={Frontiers in Pediatrics},
  volume={12},
  pages={1451466},
  year={2025},
  publisher={Frontiers Media SA}
}

@article{Qwen2.5-vl,
  title={Qwen2. 5-vl technical report},
  author={Bai, Shuai and Chen, Keqin and Liu, Xuejing and Wang, Jialin and Ge, Wenbin and Song, Sibo and Dang, Kai and Wang, Peng and Wang, Shijie and Tang, Jun and others},
  journal={arXiv preprint arXiv:2502.13923},
  year={2025}
}

@inproceedings{clip,
  title={Learning transferable visual models from natural language supervision},
  author={Radford, Alec and Kim, Jong Wook and Hallacy, Chris and Ramesh, Aditya and Goh, Gabriel and Agarwal, Sandhini and Sastry, Girish and Askell, Amanda and Mishkin, Pamela and Clark, Jack and others},
  booktitle={International conference on machine learning},
  pages={8748--8763},
  year={2021},
  organization={PmLR}
}

@article{BiomedGPT,
  title={A generalist vision--language foundation model for diverse biomedical tasks},
  author={Zhang, Kai and Zhou, Rong and Adhikarla, Eashan and Yan, Zhiling and Liu, Yixin and Yu, Jun and Liu, Zhengliang and Chen, Xun and Davison, Brian D and Ren, Hui and others},
  journal={Nature medicine},
  volume={30},
  number={11},
  pages={3129--3141},
  year={2024},
  publisher={Nature Publishing Group US New York}
}

@article{RadFM,
  title={Towards generalist foundation model for radiology by leveraging web-scale 2d\&3d medical data},
  author={Wu, Chaoyi and Zhang, Xiaoman and Zhang, Ya and Hui, Hui and Wang, Yanfeng and Xie, Weidi},
  journal={Nature Communications},
  volume={16},
  number={1},
  pages={7866},
  year={2025},
  publisher={Nature Publishing Group UK London}
}

@article{paddleocr,
  title={Paddleocr 3.0 technical report},
  author={Cui, Cheng and Sun, Ting and Lin, Manhui and Gao, Tingquan and Zhang, Yubo and Liu, Jiaxuan and Wang, Xueqing and Zhang, Zelun and Zhou, Changda and Liu, Hongen and others},
  journal={arXiv preprint arXiv:2507.05595},
  year={2025}
}

@article{zhipu,
  title={Chatglm: A family of large language models from glm-130b to glm-4 all tools},
  author={Glm, Team and Zeng, Aohan and Xu, Bin and Wang, Bowen and Zhang, Chenhui and Yin, Da and Zhang, Dan and Rojas, Diego and Feng, Guanyu and Zhao, Hanlin and others},
  journal={arXiv preprint arXiv:2406.12793},
  year={2024}
}

@article{Flamingo,
  title={Flamingo: a visual language model for few-shot learning},
  author={Alayrac, Jean-Baptiste and Donahue, Jeff and Luc, Pauline and Miech, Antoine and Barr, Iain and Hasson, Yana and Lenc, Karel and Mensch, Arthur and Millican, Katherine and Reynolds, Malcolm and others},
  journal={Advances in neural information processing systems},
  volume={35},
  pages={23716--23736},
  year={2022}
}

@inproceedings{blip2,
  title={BLIP-2: Bootstrapping Language-Image Pre-training with Frozen Image Encoders and Large Language Models},
  author={Li, Junnan and Li, Dongxu and Savarese, Silvio and Hoi, Steven C. H.},
  booktitle={International Conference on Machine Learning (ICML)},
  year={2023}
}

@article{Intern2.5-vl,
  title={Expanding Performance Boundaries of Open-Source Multimodal Models with Model, Data, and Test-Time Scaling},
  author={Zhe Chen and Weiyun Wang and Yue Cao and Yangzhou Liu and Zhangwei Gao and Erfei Cui and Jinguo Zhu and Shenglong Ye and Hao Tian and Zhaoyang Liu and Lixin Gu and Xuehui Wang and Qingyun Li and Yiming Ren and Zixuan Chen and Jiapeng Luo and Jiahao Wang and Tan Jiang and Bo Wang and Conghui He and Botian Shi and Xingcheng Zhang and Han Lv and Yi Wang and Wenqi Shao and Pei Chu and Zhongying Tu and Tong He and Zhiyong Wu and Hui Deng and Jiaye Ge and Kaiming Chen and Min Dou and Lewei Lu and Xizhou Zhu and Tong Lu and Dahu Lin and Yunfeng Qiao and Jifeng Dai and Wenhai Wang},
  journal={ArXiv},
  year={2024},
  volume={abs/2412.05271}
}

@article{Qwen3-vl,
  title={Qwen3-VL Technical Report},
  author={Shuai Bai and Yuxuan Cai and Ruizhe Chen and others},
  journal={ArXiv},
  year={2025},
  volume={abs/2511.21631}
}

@inproceedings{RougeL,
  title={Rouge: A package for automatic evaluation of summaries},
  author={Lin, Chin-Yew},
  booktitle={Text summarization branches out},
  pages={74--81},
  year={2004}
}

@inproceedings{beyond,
  title={Beyond n-grams: A hierarchical reward learning framework for clinically-aware medical report generation},
  author={Wang, Yuan and Gao, Shujian and Liu, Jiaxiang and Jiang, Songtao and Haoxiang, Xia and Zhang, Xiaotian and Kang, Zhaolu and Wang, Yemin and Liu, Zuozhu},
  booktitle={Proceedings of the AAAI Conference on Artificial Intelligence},
  volume={40},
  number={40},
  pages={33719--33727},
  year={2026}
}

@article{Ratescore,
  title={Ratescore: A metric for radiology report generation},
  author={Zhao, Weike and Wu, Chaoyi and Zhang, Xiaoman and Zhang, Ya and Wang, Yanfeng and Xie, Weidi},
  journal={arXiv preprint arXiv:2406.16845},
  year={2024}
}

@article{MIMIC-CXR,
  title={MIMIC-CXR, a de-identified publicly available database of chest radiographs with free-text reports},
  author={Johnson, Alistair EW and Pollard, Tom J and Berkowitz, Seth J and Greenbaum, Nathaniel R and Lungren, Matthew P and Deng, Chih-ying and Mark, Roger G and Horng, Steven},
  journal={Scientific data},
  volume={6},
  number={1},
  pages={317},
  year={2019},
  publisher={Nature Publishing Group UK London}
}

@article{IU-XRAY,
  title={Preparing a collection of radiology examinations for distribution and retrieval},
  author={Demner-Fushman, Dina and Kohli, Marc D and Rosenman, Marc B and Shooshan, Sonya E and Rodriguez, Laritza and Antani, Sameer and Thoma, George R and McDonald, Clement J},
  journal={Journal of the American Medical Informatics Association},
  volume={23},
  number={2},
  pages={304--310},
  year={2015},
  publisher={Oxford Academic}
}

@inproceedings{Bleu,
  title={Bleu: a method for automatic evaluation of machine translation},
  author={Papineni, Kishore and Roukos, Salim and Ward, Todd and Zhu, Wei-Jing},
  booktitle={Proceedings of the 40th annual meeting of the Association for Computational Linguistics},
  pages={311--318},
  year={2002}
}

@inproceedings{loss,
  title={Enhanced contrastive learning with multi-view longitudinal data for chest x-ray report generation},
  author={Liu, Kang and Ma, Zhuoqi and Kang, Xiaolu and Li, Yunan and Xie, Kun and Jiao, Zhicheng and Miao, Qiguang},
  booktitle={Proceedings of the Computer Vision and Pattern Recognition Conference},
  pages={10348--10359},
  year={2025}
}

@article{Llava-onevision,
  title={Llava-onevision: Easy visual task transfer},
  author={Li, Bo and Zhang, Yuanhan and Guo, Dong and Zhang, Renrui and Li, Feng and Zhang, Hao and Zhang, Kaichen and Zhang, Peiyuan and Li, Yanwei and Liu, Ziwei and others},
  journal={arXiv preprint arXiv:2408.03326},
  year={2024}
}

@inproceedings{METEOR,
  title={METEOR: An automatic metric for MT evaluation with improved correlation with human judgments},
  author={Banerjee, Satanjeev and Lavie, Alon},
  booktitle={Proceedings of the acl workshop on intrinsic and extrinsic evaluation measures for machine translation and/or summarization},
  pages={65--72},
  year={2005}
}


\end{document}